\documentclass{article}

\usepackage[utf8]{inputenc}
\usepackage{textcomp}
\usepackage{microtype}
\usepackage{graphicx}
\usepackage{subfigure}
\usepackage{booktabs} 
\usepackage{soul}
\usepackage{enumitem}
\usepackage{amsmath}
\usepackage{makecell}
\usepackage{multirow}
\usepackage{vwcol}
\usepackage{siunitx}
\usepackage[table,x11names]{xcolor}
\definecolor{lightergray}{gray}{0.93}

\DeclareUnicodeCharacter{03BC}{\TextOrMath{\textmu}{\mu}}

\usepackage{hyperref}

\setitemize{topsep=0pt, itemsep=0pt, parsep=2pt}
\setenumerate{topsep=0pt, itemsep=0pt, parsep=2pt}

\usepackage[accepted]{mlsys2020}

\mlsystitlerunning{μNAS: Constrained Neural Architecture Search for Microcontrollers}

\begin{document}

\newif\ifpreprint
\preprinttrue

\ifpreprint
\newcommand{\releaseurl}{\url{https://github.com/eliberis/uNAS}}
\makeatletter
\renewcommand{\mlsys@appearing}{}
\makeatother
\else
\newcommand{\releaseurl}{\textit{(In progress. URL to be provided after review.)}}
\fi

\twocolumn[
\mlsystitle{μNAS: Constrained Neural Architecture Search\\ for Microcontrollers}




\begin{mlsysauthorlist}
\mlsysauthor{Edgar Liberis}{cam}
\mlsysauthor{Łukasz Dudziak}{sam}
\mlsysauthor{Nicholas D. Lane}{cam,sam}
\end{mlsysauthorlist}

\mlsysaffiliation{cam}{Department of Computer Science and Technology, University of Cambridge, United Kingdom}
\mlsysaffiliation{sam}{Samsung AI Centre Cambridge, Cambridge, United Kingdom}

\mlsyscorrespondingauthor{Edgar Liberis}{el398 \textit{at} cam.ac.uk}

\mlsyskeywords{Machine Learning, MLSys, MCU, NAS, Deep learning, Neural Networks, Model Compression, Architecture Search, Microcontrollers, TinyML}

\vskip 0.3in

\begin{abstract}
IoT devices are powered by microcontroller units (MCUs) which are extremely resource-scarce: a typical MCU may have an underpowered processor and around 64 KB of memory and persistent storage, which is orders of magnitude fewer computational resources than is typically required for deep learning. Designing neural networks for such a platform requires an intricate balance between keeping high predictive performance (accuracy) while achieving low memory and storage usage and inference latency. This is extremely challenging to achieve manually, so in this work, we build a neural architecture search (NAS) system, called μNAS, to automate the design of such small-yet-powerful MCU-level networks. μNAS explicitly targets the three primary aspects of resource scarcity of MCUs: the size of RAM, persistent storage and processor speed. μNAS represents a significant advance in resource-efficient models, especially for “mid-tier” MCUs with memory requirements ranging from 0.5 KB to 64 KB. We show that on a variety of image classification datasets μNAS is able to (a) improve top-1 classification accuracy by up to 4.8\%, or (b) reduce memory footprint by 4--13${\times}$, or (c) reduce the number of multiply-accumulate operations by at least 2${\times}$, compared to existing MCU specialist literature and resource-efficient models. μNAS is freely available for download at \releaseurl
\end{abstract}
]



\printAffiliationsAndNotice{}  

\section{Introduction}
\label{introduction}

We would like to use deep learning to add computational intelligence to small personal IoT devices. Running neural networks on-device would allow a higher degree of privacy and autonomy, due to the computation happening locally and the user's data never leaving the device. However, deep learning typically requires much more computational power than is available on these devices, forcing them to rely on a remote compute server or a companion smartphone. 

IoT devices are powered by microcontroller units (MCUs). MCUs are ultra-small computers with a low-frequency processor, a persistent program memory (Flash) and volatile static RAM---all contained within a single chip. They are orders of magnitude cheaper and more power-efficient compared to mobile phones and desktops, which contributes to their widespread usage: more than 30B units were estimated to have shipped in 2019~\cite{mcumarket2020}. However, these benefits come at the cost of drastically reduced computational power. Here, we consider ``mid-tier'' IoT-sized MCUs with up to 64KB of SRAM and 64KB of persistent storage available\footnote{More powerful ``top-tier'' alternatives are also available at a higher cost, such as ARM Cortex M4- or M7-based MCUs (up to ${\approx}$400 MHz clock frequency, $\leq$ 512KB SRAM, $\leq$ 1MB Flash). Considering “mid-tier” MCUs allows our methodology to be applied more broadly both now and also in the future, once the current “mid-tier” becomes the new “low-” and the most widespread tier.} for performing neural network inference. These severe resource constraints present a challenge to running resource-intense computations.

Designing neural networks with small computational requirements is tackled by the field of model compression. Methods such as pruning, distillation, quantisation and cheaper layers, have been developed to obtain compressed models from large networks in a way that tries to minimise the amount of lost accuracy~\cite{cheng2017survey}. This also reveals a trade-off between the accuracy of a neural network and its resource usage, or size, suggesting it is difficult to have small models that generalise well. 

However, most model compression methods are not suitable for producing MCU-sized neural networks. Many techniques target mobile devices (or similar platforms), which have orders of magnitude of more computational resources than MCUs, or focus on reducing the number of parameters (model size) or floating-point operations (FLOPs) within the network's layers while keeping the overall architecture (layer connectivity) intact, which does not fully capture all types of resource scarcity of an MCU. Even manually designed resource-efficient models, such as MobileNet and SqueezeNet, \textit{would exceed the assumed resource budget by over 10${\times}$}. This necessitates further research into specialist methods for deep learning on MCUs.

To illustrate the deployment challenges, let us briefly go through design considerations of executing a neural network on an MCU and how the resource scarcity affects it: 
\begin{itemize}
\item Temporary data generated by the neural network (activation matrices) need to fit within the SRAM of the MCU ($<$ 64 KB). 
\item Any static data, such as the neural network's parameters and program code, need to fit within the ROM / Flash memory of the MCU ($<$ 64 KB). 
\item A network needs to execute quickly to keep the inference process within reasonable power and time constraints while running on an underpowered processor.
\end{itemize}

To tackle neural network design under hard constraints, we turn to neural architecture search (NAS). NAS is related to model selection and hyperparameter tuning but refers specifically to selecting an architecture from a large predefined space of all neural networks, called \textit{the search space}. During the search, many common NAS algorithms consider numerous candidate models, each assembled from a predefined set of building blocks, to find the one with the highest accuracy~\cite{elsken2019neural}. When properly conditioned, NAS can produce architectures within certain constraints or targeting several objectives at once, though this also increases the difficulty of the search problem.

While NAS is a promising direction to consider when designing models for MCUs, most current NAS systems target much larger platforms, such as GPUs or mobile devices. We would expect them to fail to produce MCU-compatible architectures because they were not designed to handle extreme resource scarcity of MCUs, owing at least to either an inability to represent MCU-sized architectures in the search space at all or doing so too coarsely, resulting in too few candidates to choose from during search.

Our work, μNAS, is one of the first few neural architecture search systems that target microcontrollers explicitly. μNAS accurately captures the resource requirements and, combined with model compression, successfully finds fast high-accuracy microcontroller-sized neural networks with a low memory footprint ($<$ 64 KB). 

In contrast to other NAS systems for MCUs~\cite{fedorov2019sparse, lin2020mcunet}, μNAS uses a standard neural network execution runtime, a fine-grained search space and accurate objective functions, which allows it to improve upon other methods on five image classification datasets using comparable MCU resource requirements. We found that μNAS can either (a) improve top-1 classification accuracy by up to 4.8\%, or (b) reduce memory usage by 4--13${\times}$, or (c) reduce the number of multiply-accumulate operations by at least $2{\times}$, depending on the task.

The main contributions of this work are:
\begin{itemize}
\item \textit{We propose and motivate a multiobjective constrained NAS algorithm suitable for finding MCU-level architectures, called μNAS.} It is assembled out of: 
\begin{enumerate}
\item a granular search space; 
\item a set of constraints that accurately capture resource scarcity of microcontroller platforms: peak memory usage, model size and latency; 
\item a search algorithm capable of optimising for multiple objectives in the said search space; 
\item network pruning, to obtain small accurate models.
\end{enumerate}
\item \textit{We perform ablation studies to quantitatively justify design decisions made in μNAS}, namely:
\begin{enumerate}
\item whether using network pruning helps find smaller models than otherwise; 
\item which search algorithm should be used;
\item how including or excluding individual objectives influences the properties of the models found.
\end{enumerate}
\item \textit{We conduct extensive experiments} over five microcontroller-friendly image classification tasks and existing literature for resource-efficient neural network to demonstrate the superior performance of μNAS.
\end{itemize}

\section{Related work}

Until recently, due to high computational and memory requirements, neural networks have not been widely used in machine learning for IoT-sized devices. Initially, elements of gradient-based learning were combined with other machine learning algorithms to produce a resource-efficient solution. ProtoNN~\cite{gupta2017protonn} learn a sparse projection matrix, and use learned class prototypes for classification. Bonsai~\cite{kumar2017resource} learn a sparse, shallow decision tree with feature extraction occurring at a decision path. Both methods have a low memory footprint of under 2KB. More recently, manually designed neural networks with quantisation and binarisation have been used for image classification on MCUs, too, though with a larger memory footprint~\cite{zhang2017hello, mocerino2019coopnet}. 

Neural architecture search (NAS) is a widely explored topic in deep learning for desktop and server GPUs. There are approaches based on reinforcement learning~\cite{zoph2016neural, zhou2018rena, tan2019mnasnet}, evolutionary algorithms~\cite{real2019regularized}, Bayesian optimisation~\cite{kandasamy2018neural,jin2019auto} and gradient optimisation~\cite{liu2018darts,mei2019atomnas}, that sometimes employ weight sharing to amortise the cost of training across multiple candidate models~\cite{pmlr-v80-pham18a, guo2019single}. Constrained multiobjective NAS has been traditionally used to find ``mobile''-level networks, such as by optimising energy~\cite{hsu2018monas} and latency~\cite{fernandez2020searching, cai2018proxylessnas} in addition to accuracy. However, few works consider MCUs as the target platform.

The closest work to ours is ``\emph{SpArSe}''~\cite{fedorov2019sparse}, which finds both sparse and dense convolutional neural networks suitable for MCUs. In many ways, this work has served us as a promising direction to take: we share similarities with multiobjective optimisation and the use of network pruning. In comparison, we: (1) improve the key aspects of search objectives, making them more faithful to how neural networks are executed on an MCU; (2) improve on the search space, allowing for more layer connectivity; and (3) we adopt an alternative search procedure and a pruning method--all of the which allows us to produce architectures with higher accuracy and lesser resource usage. 

Recently, \citet{lin2020mcunet} developed a neural network execution runtime for MCUs together with a NAS tailored to it (for comparison, we use an off-the-shelf runtime described in Section~\ref{sec:mcu_runtime}). To the best of our understanding, the runtime uses partial operator evaluation to store only one or more columns\footnote{Convolution can be implemented by an \texttt{im2col} transposition followed by matrix multiplication. Thus here, a ``column'' is a window of the input covered by a convolutional kernel.} of the output matrix of each operator. The NAS targets larger MCUs, of $>$ 256 KB SRAM and Flash, and performs the search by selecting a subnetwork from a larger model. This limits the connectivity of resulting models (\textit{i.e.} fewer architectures are considered during search) but allows for faster convergence and targeting a more difficult task of ImageNet classification. We compare discovered models for the Speech Commands dataset: we find that μNAS finds models with a smaller memory footprint (in our range of interest) without using a custom runtime.

\ifpreprint
Orthogonally to network design, optimising compilers for deep learning are being developed to ease the deployment of neural networks onto an MCU. These methods focus on optimising the execution of a given neural network, rather than searching for optimal neural network architecture. For example, µTVM~\cite{mutvm2020} optimises the execution by searching for an efficient implementation of a model's layers (\emph{kernels}, such as convolution or matrix multiplication), performs scheduling and code generation for a specified target microcontroller device.
\fi

\section{Design of μNAS}
\label{sec:design}

We believe the following two design requirements are essential for an MCU-level NAS and set it apart from mobile- or GPU-level NAS systems:
\begin{enumerate}
    \item \textbf{A highly granular search space.} To discover accurate networks that fit within the strict resource requirements, NAS needs to control all aspects of a network: its layers, their size (such as output size or convolutional kernel size) and connectivity. In contrast, NAS for GPUs typically uses search spaces with a larger resolution, \emph{e.g.} manipulating groups of layers at a time or replicating layers in predefined patterns at predetermined input resolutions. (The granularity of the search space is further discussed in Sections~\ref{sec:search_space} and \ref{sec:search_algorithm}.) 
    \item \textbf{Accurate resource use computation.} To know which models fit within resource constraints, NAS needs to efficiently and accurately calculate or simulate how each candidate network will use memory, storage and computational resources (based on an assumed execution strategy). A precise computation is not often needed for models running on GPUs or mobiles, as these platforms are not as resource-constrained as MCUs.
\end{enumerate}

In the remainder of this section, we discuss the motivation behind the requirements above and how they are implemented in μNAS. We start by describing how a neural network is executed on an MCU and, after briefly formalising the problem of multiobjective NAS, we will talk about major components in μNAS, namely the search space, objective functions, search algorithms and compression, that are all crucial for finding performant MCU-sized architectures.

\subsection{Neural network execution on MCUs}
\label{sec:mcu_runtime}
A model's resource usage depends on the runtime, that is on how the software chooses to run the network for inference and manage memory. Knowledge of this has to be incorporated into the NAS to correctly estimate a candidate model's resource usage during the search. Here, we follow the execution strategy used by the TensorFlow Lite Micro~\cite{david2020tensorflow} interpreter and CMSIS-NN libraries:
\begin{enumerate}
    \item Operators (neural network layers) are executed in a predefined order, one at a time (no parallelism).
    \item An operator is executed by (a) allocating/reserving memory for its output buffer in SRAM, (b) fully executing the body of an operator and (c) deallocating its input buffers (if not used elsewhere later on). Hence, the activation matrices of a neural network are only stored within SRAM.
    \item Any static data, such as neural network weights, are read from the executable binary, stored in Flash.
    \item The network's weights and all computation are quantised to an 8-bit fixed precision data type (\texttt{int8}), \emph{e.g.} using the affine quantisation~\cite{jacob2018quantization}. 
    \item No operators are executed partially, or more than once, and no data is written to external writeable storage (an SD-card), as this would be prohibitively slow. 
\end{enumerate}

It is also possible to adopt a custom execution framework that, for example, leverages partial execution, or operator fusion, and adapt μNAS for that instead. However, by using a widely adopted and supported framework, μNAS can deliver state-of-the-art results that can be readily used, without having to maintain a custom runtime. 

\subsection{What is Neural Architecture Search (NAS)?}
At its core, NAS is a constrained zeroth-order optimisation problem, where we seek to find a neural network $\alpha^*$ (belonging to the search space $\mathcal{A}$) that maximises some objective function (or goal), called $\mathcal{L}$, such as the accuracy on the target dataset with respect to some resource usage constraints. We assume that the validation set accuracy for a particular architecture $\alpha$ is maximised by some weights $\theta^*$, obtained via gradient descent. Evaluating $\mathcal{L}(\alpha)$ is expensive since it requires training a model, so we seek to find the optimum $\alpha^*$ in a limited number of queries to $\mathcal{L}$ (evaluations of $\mathcal{L}$). 

Resource constraints can be treated as soft constraints, \emph{i.e.} reformulated as extra objectives with penalty terms. This turns NAS into a multiobjective optimisation problem (rewritten as a minimisation problem below). We include four objectives, three of which are resource constraints (discussed in Section~\ref{sec:constraints}): (1) top-1 validation set accuracy, (2) peak memory usage, (3) model size and (4) latency. 
\begin{equation}
\begin{aligned}
& \:\alpha^* & =\: &\text{argmin}_{\alpha \in \mathcal{A}} && \mathcal{L}(\alpha) \\
&         & =\: &\text{argmin}_{\alpha \in \mathcal{A}} &\{ & 1.0 - \textsc{ValAccuracy}(\alpha), \\
&         & & & &\textsc{ModelSize}(\alpha), \\
&         & & & &\textsc{PeakMemUsage}(\alpha), \\
&         & & & &\textsc{Latency}(\alpha)\} 
\end{aligned}
\end{equation}
A solution to a multiobjective problem is no longer a singular value. It is common to consider a Pareto front as a set of potential solutions: a set of points on which one objective function cannot be improved without making another objective worse. In Section~\ref{sec:search_algorithm}, we discuss a \textit{scalarisation} approach that turns the multiobjective goal into a single objective that still encourages exploring the Pareto front.

\ifpreprint
We will now discuss how to design (a) the search space $\mathcal{A}$, (b) the objective functions, (c) the algorithm that can successfully explore the search space, handle multiple objectives and explore the Pareto front, and (d) the training procedure for the candidate networks.
\fi

\subsection{Search space}
\label{sec:search_space}
Here, neural networks are directed acyclic graphs (DAGs), with nodes and edges representing operators (layers) and their connectivity. For convolutional neural networks (CNNs), operator options include convolution (at different kernel sizes, numbers of channels or strides), pooling, addition and matrix multiplication (fully-connected layers). 

State-of-the-art GPU-level CNNs can contain a large number of operators, \emph{e.g.} $>$ 100 layers~\cite{he2016deep}. If architecture search were to be used to design a comparably large model while having full freedom over layer connectivity, the problem would become intractable due to the size of the search space. Instead, some NASes constrain the space to 1 or 2 small ``cells'' (micro-architectures) to curb the number of connectivity options~\cite{liu2018darts}. These cells are then replicated in a predefined pattern to produce the final architecture (macro-architecture). Alternatively, the search space can contain pruned versions of some predefined large super-architecture~\cite{guo2019single, lin2020mcunet}. 

However, NAS is known to find good performing architectures that are wired in unexpected ways~\cite{cheng2019swiftnet}. So to give μNAS more freedom when searching under already tight resource requirements, we avoid imposing big structural constraints. As one would not expect to run large models on an MCU (they would violate the resource constraints), allowing more options in layer connectivity here does not make the search problem intractable.

\newcommand\sbullet[1][.5]{\mathbin{\vcenter{\hbox{\scalebox{#1}{$\bullet$}}}}}
\newcommand{\ipoint}{$\;\sbullet\;$}
\newcommand{\iipoint}{$\quad\sbullet\;$}
\begin{table*}[t]
    \centering
    \begin{tabular}[t]{p{9cm}ll}
    \toprule
    \textbf{Degree of freedom} & \textbf{Options} & \textbf{Morphisms} \\
    \midrule
    An architecture is $N$ ``convolutional'' blocks, where each: & $N$ in [1; 10] & append or remove random \\
    \ipoint connects either in series or in parallel to the previous one & \textit{\{parallel, serial\}} & change one to other \\
    \ipoint has $M$ convolution layers, where each layer: & $M$ in [1; 3] & 
insert or remove random \\
    \iipoint optionally, has a preceding $2{\times}2$ max-pooling operation; & \textit{\{yes, no\}} & change one to other \\
    \vspace{-26.5pt}\iipoint is either a full or a depthwise convolution with stride $S$, $C$ channels $K{\times}K$ kernel size ($S$ = 1 for $1{\times}1$ convolution and $C$ is not configured for depthwise convolution) & \makecell[l]{\textit{\{full, d/wise\}} \\
$K$ in \{1, 3, 5, 7\} \\
$C$ in [1; 128] \\
$S$ in \{1, 2\}} &
\makecell[l]{change types \\
change K by $\pm$ 2 \\
change C by $\pm$ 1, 3, 5 \\
change S by $\pm$ 1} \\
    \iipoint is optionally followed by batch norm.; & \textit{\{yes, no\}} & change one to other \\
    \iipoint is optionally followed by ReLU; & \textit{\{yes, no\}} & change one to other \\
    
    followed by a $P{\times}P$ average or maximum pooling, & \makecell[l]{
    \textit{\{avg, max\}} \\ $P$ in \{2, 4, 6\}
    } & \makecell[l] {
    change one to other \\
    change $P$ by $\pm$ 2
    } \\
    followed by $F$ fully-connected layers, where each: & $F$ in [1; 3] & 
insert or remove random \\
    \ipoint has $U$ units (output dimension), followed by a ReLU & $U$ in [10; 256] & 
change $U$ by $\pm$ 1, 3, 5 \\
followed by a final fully-connected layer ($U$ = number of classes). \\
\bottomrule
    \end{tabular}
    \caption{A template for candidate models, with free variables, their morphisms and bounds, which define the search space of μNAS.}
    \label{tab:search_space}
\end{table*}

MCU-level architectures are very sensitive to layer hyperparameters, such as the numbers of channels or units in convolutional and fully-connected layers. For example, choosing between a conv. layer with 172 and 192 channels is unlikely to make a meaningful difference for a GPU-sized model (though the former may have lower accuracy), but on an MCU choosing the larger layer may tip the model over the strict memory budget. This requires considering hyperparameters at a high granularity, which is not commonly needed for GPU-level NAS and has the negative effect of enlarging the search space. 

Thus we conclude that a sensible MCU search space would consist of small models, with few restrictions on layer connectivity and highly granular hyperparameter options. The search algorithms (described in Section~\ref{sec:search_algorithm}) navigate the search space by generating random architectures and applying changes to them (\textit{morphisms}) to produce derivative (child) networks. Morphisms here only affect the architecture: we do not attempt to preserve and transform learned weights between the parent and the child networks.

Our search space, together with all degrees of freedom, parameter limits and morphisms, is given in Table~\ref{tab:search_space}. This highly granular search space comprises $1.15 \times 10^{152}$ models.

\subsection{Resource constraints}
\label{sec:constraints}
We focus on three constraints to be used in the search, each representing a key aspect of resource scarcity on MCUs: peak memory usage, model size and latency.

\textbf{Peak memory usage.}
Under the execution strategy described earlier, an operator's input and output buffers have to be present in memory during its execution. Additionally, depending on the architecture of the network, there can be other buffers in memory that will be required by subsequent operators and thus cannot be deallocated yet. Overall, let us call the set of tensors that need to be present in memory at each step the \emph{working set}. 

As execution proceeds, tensors are allocated and deallocated at different times, changing the working set. To execute the model on an MCU, \emph{the memory occupied by the working set at its \underline{peak} must be lower than the amount of SRAM}.   

\begin{figure}[ht]
    \centering
    \includegraphics[scale=0.85]{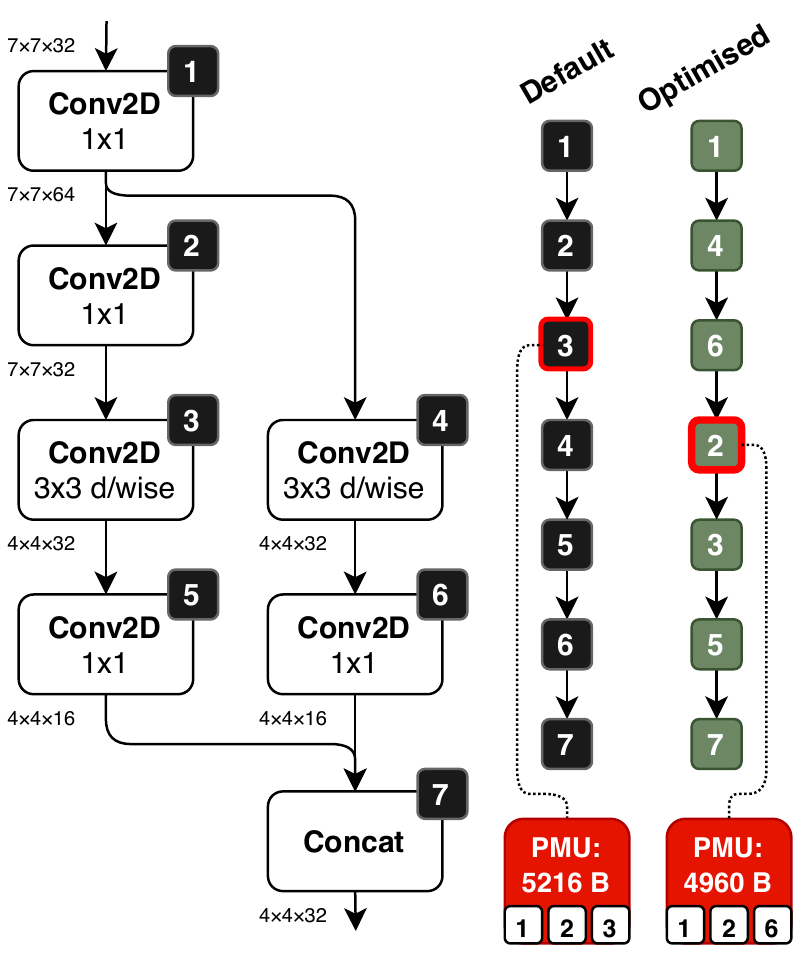}
    \caption{An example computation graph where the default and optimised execution paths yield different peak memory usage (PMU)}
    \label{fig:example_pmu}
\end{figure}

Perhaps surprisingly, it is not straightforward to compute the peak memory usage of a network when it has branches. Branches are commonly featured in popular CNNs as residual connections~\cite{he2016deep}, Inception modules~\cite{szegedy2015going}, or NAS-designed cells~\cite{cheng2019swiftnet}. Here's why: upon reaching a branching point, the inference framework has a choice which operator to execute next. Different execution orders change which tensors constitute the working set, which in turn affects the peak memory usage (see example in Figure~\ref{fig:example_pmu}). Thus, to most accurately capture the minimum amount of memory required to run a network, a NAS has to compute an execution order that gives the smallest peak working set size. 

To achieve this, we build upon the algorithm by \citet{liberis2019neural}, which enumerates all topological orders of a network's computational graph to find one that yields the minimal peak memory usage. We extend it to model other features, such as reusing input buffers of an `Add' operator. 

To the best of our knowledge, this is the first work to compute memory usage accurately during the search, without relying on on-device benchmarking (that is often slow) or using under-approximations. Later (Section~\ref{sec:search_algorithm}), we will discuss that it is reasonable to assume that the most performant networks will have high resource usage. This makes a precise memory usage computation essential in identifying which candidate networks lie just below the resource limits.

\textbf{Model size.}
All static data, such as code and parameters (weights) of a neural network, are stored in the persistent (Flash) memory of an MCU. Traditionally, each parameter is represented by a 32-bit floating-point number (a \texttt{float}), which occupies 4 bytes. However, we can reduce the per-parameter storage requirement by using quantisation. 

We assume an integer-only quantisation technique used in TensorFlow Lite, which quantises each parameter from 32-bit floats to 8-bit integers after training (by calibrating quantisation parameters on the validation set). This is an excellent choice for microcontrollers, as it is: (a) byte-aligned (thus a value can be loaded directly with no decoding), (b) does not require any floating-point arithmetic units to be present on the chip, (c) is also widely adopted and (d) does not make networks suffer from accuracy loss during quantisation (confirmed by our and MCUNet~\cite{lin2020mcunet} experiments). If needed, μNAS can be easily adapted for other quantisation techniques. 

Thus to compute the storage requirement, NAS simply counts the number of parameters of a neural network at 8 bits = 1 byte per parameter.

\textbf{Latency.}
In order to discover models that run sufficiently quickly on an MCU, NAS has to have a notion of how long it takes to perform a single inference: \textit{model latency}. Some NAS works have estimated latency by either using a proxy metric, such as the number of floating-point operations (FLOPs) required to complete a forward pass~\cite{mei2019atomnas, xie2018snas} or predicting the latency via a surrogate model. The latter can be either a sum of latency predictions of each layer or a prediction for a whole model which takes into account inter-layer interactions~\cite{chau2020brp, cai2018proxylessnas}.

Using a predictive model has become the dominant approach for GPU-/mobile-level NAS~\cite{tan2019mnasnet, chau2020brp}, as proxy metrics, such as FLOPs, fail to account for scheduling, caching, parallelism and other properties of the inference software or hardware. However, MCUs typically lack these performance-enhancing features: the software runs on a single-core processor at a fixed frequency with no data caching, which makes modelling the runtime simpler.

\begin{figure}[t]
    \centering
    \includegraphics[scale=0.8]{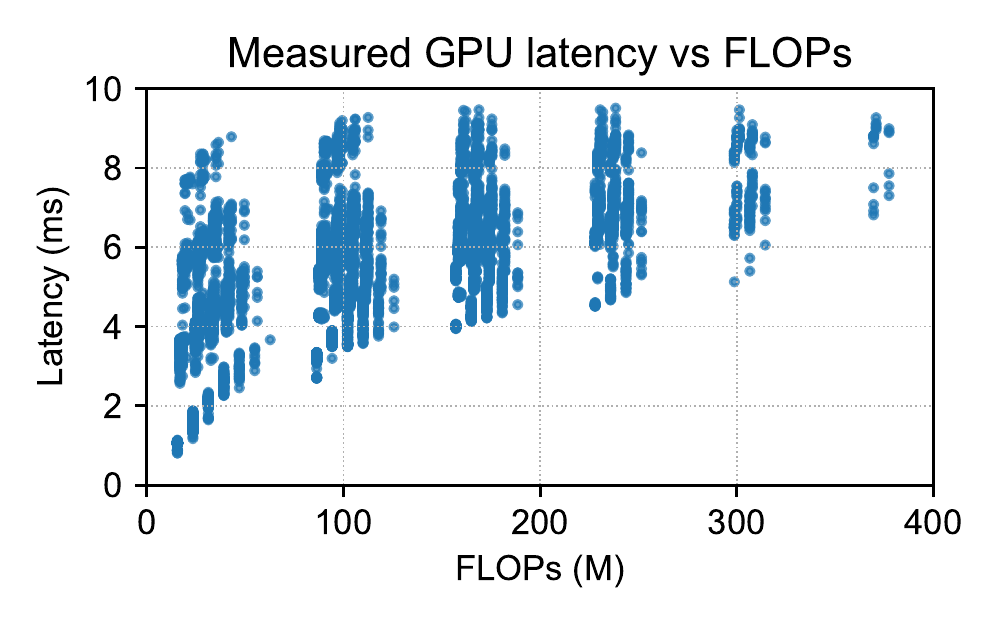}\\
    \includegraphics[scale=0.8]{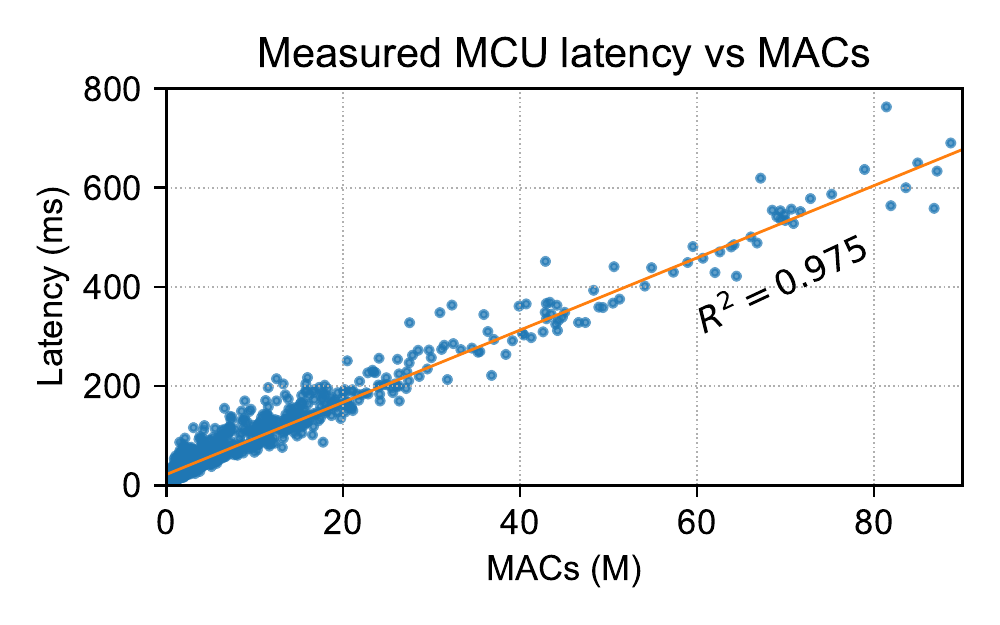}
    \caption{\textit{(Top)} A typical plot of FLOPs vs model latency on a desktop GPU, reproduced from \citet{chau2020brp}. \textit{(Bottom)} Our measured MACs vs model latency on a sample of 1000 models from our search space. The models ran on a NUCLEO-H743ZI2~\cite{nucleoh743zi2020} MCU board using the TensorFlow Lite Micro runtime. The values were averaged over ten runs. \textbf{The data shows that MACs are a good predictor of model latency on MCUs.}}
    \label{fig:mcu_latency}
\end{figure}

We settle on using a number of multiply-accumulate operations (MACs) as a proxy for model latency. To verify that this estimator approximates actual model latency well, in Figure~\ref{fig:mcu_latency}, we plot the measured runtime of a 1000 random models from our search space and versus the number of MAC operations. For reference, we reproduce the figure from BRP-NAS~\cite{chau2020brp} to show how the number of FLOPs compares to latency on a desktop GPU. We observe that MACs are \emph{not} systematically under- or over-approximating latency on an MCU, and the result has an $R^2 = 0.975$ goodness-of-fit.

\subsection{Search algorithms}
\label{sec:search_algorithm}
Intuitively, because performant neural networks are large, we would expect the best architectures to make use of all available resources and thus be located at a boundary between models that fit within the resource constraints and ones that do not. By design of the search space (both ours and others), a change in a single axis (degree of freedom) can significantly change layer properties and inter-connectivity, thus potentially drastically altering the model’s resource requirements and accuracy. This makes the boundary jagged, rendering multiobjective NAS a challenging task for black-box optimisation methods. 

Now, we establish a way to handle multiple objectives in a single goal and discuss two optimisation algorithms implemented in μNAS: aging evolution (AE) and Bayesian optimisation (BO). Despite being inherently sequential, both AE and BO allow for parallel point evaluation~\cite{real2019regularized, kandasamy2018parallelised}, allowing us to scale the search over multiple GPUs. We compare the two in our experiments to discover the best performing approach.

\textbf{Handling multiple objectives.}
Akin to \textit{SpArSe}, we use random scalarisations (\citealp{paria2020flexible}, TS expression) to combine multiple objectives into a single goal (the optimisation target function).
\begin{equation}\label{eqn:scalarised_goal}
\mathcal{L}^t(\alpha) = \max
\begin{cases}
  \lambda^t_1 \ (1.0 - \textsc{ValAccuracy}(\alpha)), \\   
  \lambda^t_2 \ \textsc{PeakMemUsage}(\alpha), \\    
  \lambda^t_3 \ \textsc{ModelSize}(\alpha), \\
  \lambda^t_4 \ \textsc{MACs}(\alpha)
\end{cases}
\end{equation}
Each objective has an associated scalar term $\lambda_i^t$, which specifies its relative importance in the current search goal. To encourage the exploration of the Pareto front, the goal changes at every search round (indexed by $t$) by resampling coefficients $\boldsymbol{\lambda}^t$. The trade-off preferences between multiple objectives can be encoded in the distribution for $\boldsymbol{\lambda}$: we use $1/\lambda_i \sim \text{Uniform}[0; b] $, where $b$ is a user-specified soft upper bound for the $i^{\text{th}}$ objective. This objective encourages exploring a narrow region of the Pareto front, where models lie within requested resource usage bounds.

\textbf{NAS via local search.}
Local search optimises the goal function by repeatedly evolving a set of candidate points. We use aging evolution (AE) as a local search algorithm for NAS~\cite{real2019regularized}. AE operates by keeping a population of $P$ architectures and, at each search round, subsampling the population to get $S$ architectures to choose the one that gives the smallest value of $\mathcal{L}^t(\alpha)$. A random morphism is then applied to this winning architecture to produce an offspring, which is then evaluated and added to the population, replacing the oldest architecture. Aging evolution has proved itself a competitive search algorithm for NAS, beating many baselines and random search.

\textbf{NAS via Bayesian optimisation.}
Bayesian optimisation (BO) uses surrogate models to approximate the target function and guide the search towards promising unexplored regions of its domain. BO is successfully used in hyperparameter optimisation~\cite{falkner2018bohb} and NAS~\cite{jin2019auto}, often using Gaussian Processes (GPs) as a surrogate model for the accuracy of a neural network.

In μNAS, at each step of the search, BO chooses next architecture by optimising the target function ($\mathcal{L}^t$), with the only difference that $\textsc{ValAccuracy}(\theta)$ is computed by sampling from the GP that approximates it (other objectives are cheap to compute and thus do not require a surrogate model, unlike in \textit{SpArSe}). GPs require a kernel that captures the notion of similarity between any two neural networks. For this, we build upon NASBOT~\cite{kandasamy2018neural}, which uses optimal transport to define a distance function between two computational graphs of neural networks; we update the kernel for our search space.

\subsection{Model compression}
\label{sec:model_compression}
\ifpreprint
Having established all essential components of μNAS, we take a look at model compression techniques, to see if incorporating them can help NAS discover small architectures with better accuracy.
\fi

Model compression methods attempt to create or train a smaller model using a bigger more performant model, while retaining as much of the predictive performance as possible. Parameter-efficient convolutional blocks, such as inverted residual blocks of MobileNet V2~\cite{sandler2018mobilenetv2}, and fire modules of SqueezeNet~\cite{iandola2016squeezenet}, are already representable in the search space, allowing μNAS to recover them if necessary.

μNAS supports using model compression during the search: in particular, we hypothesise that network pruning would help in finding small architectures with high accuracy---we will empirically determine this in our experiments.

\textbf{Pruning.} Pruning removes individual parameters from a neural network that do not significantly affect generalisation. We use structured pruning, which eliminates entire groups of parameters: channels (in conv. layers) or units/neurons (in fully-connected layers), resulting in smaller dense models, as opposed to sparsifying the weight matrices. This makes the search and the pruning share the task of determining the model’s hyperparameters: the search produces a base network, which is then adjusted by pruning in a more informed way by discarding channels/units that were deemed unimportant during training.

μNAS employs DPF pruning~\cite{lin2020dynamic} which uses the $L_2$ norm of channels/units to discard weight groups until the desired proportion of groups is removed. μNAS sets this target ``sparsity'' proportion and the network is gradually pruned during training until the target proportion is reached.

\subsection{Summary of the μNAS search procedure}
\label{sec:summary}
As an example, let us walk through the search procedure when μNAS is used with aging evolution (AE) and pruning:
\begin{enumerate}
    \item \textbf{AE initialisation.} A random initial population of networks is generated and trained, together with their target “sparsity” values (random in the allowed range). 
    \item \textbf{Preamble.} A new search step $t$ begins by generating a new goal function. This is done by resampling coefficients $\mathbf{\lambda}$ in Equation~\ref{eqn:scalarised_goal}.
    \item \textbf{Search algorithm body.} The search proceeds by the rules of AE: the next architecture to be trained is determined by applying a random morphism (Table~\ref{tab:search_space}) to the chosen parent. The target “sparsity” level for the new model is inherited and randomly perturbed.
    \item \textbf{Write back.} Once the training and pruning are complete, the values of all four objectives of the pruned network (the final accuracy, peak memory usage, storage usage and the number of MACs) are recorded, and the network is added to the population. A new round starts from step 2.
\end{enumerate}

\section{Evaluation}
\label{sec:evaluation}

\begin{figure*}[t]
    \centering
    \includegraphics[scale=0.60]{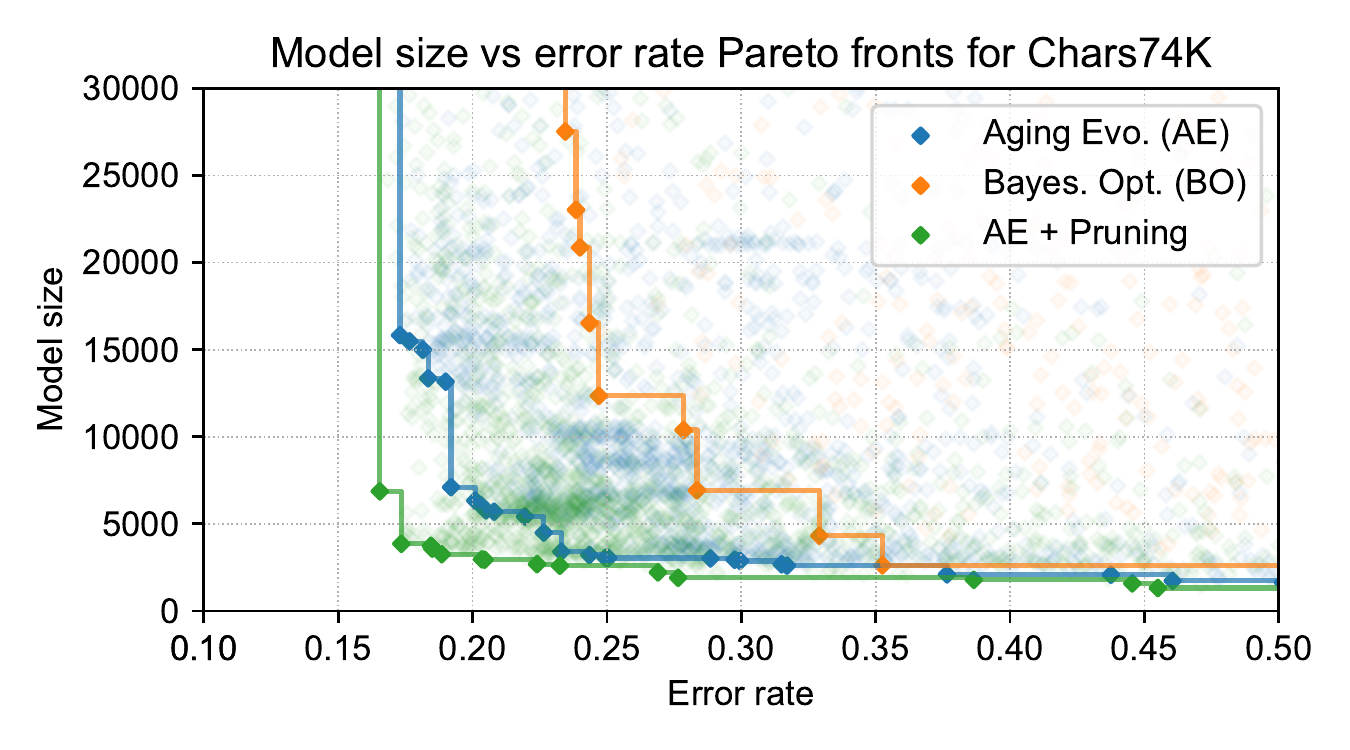}%
    \quad%
    \includegraphics[scale=0.60]{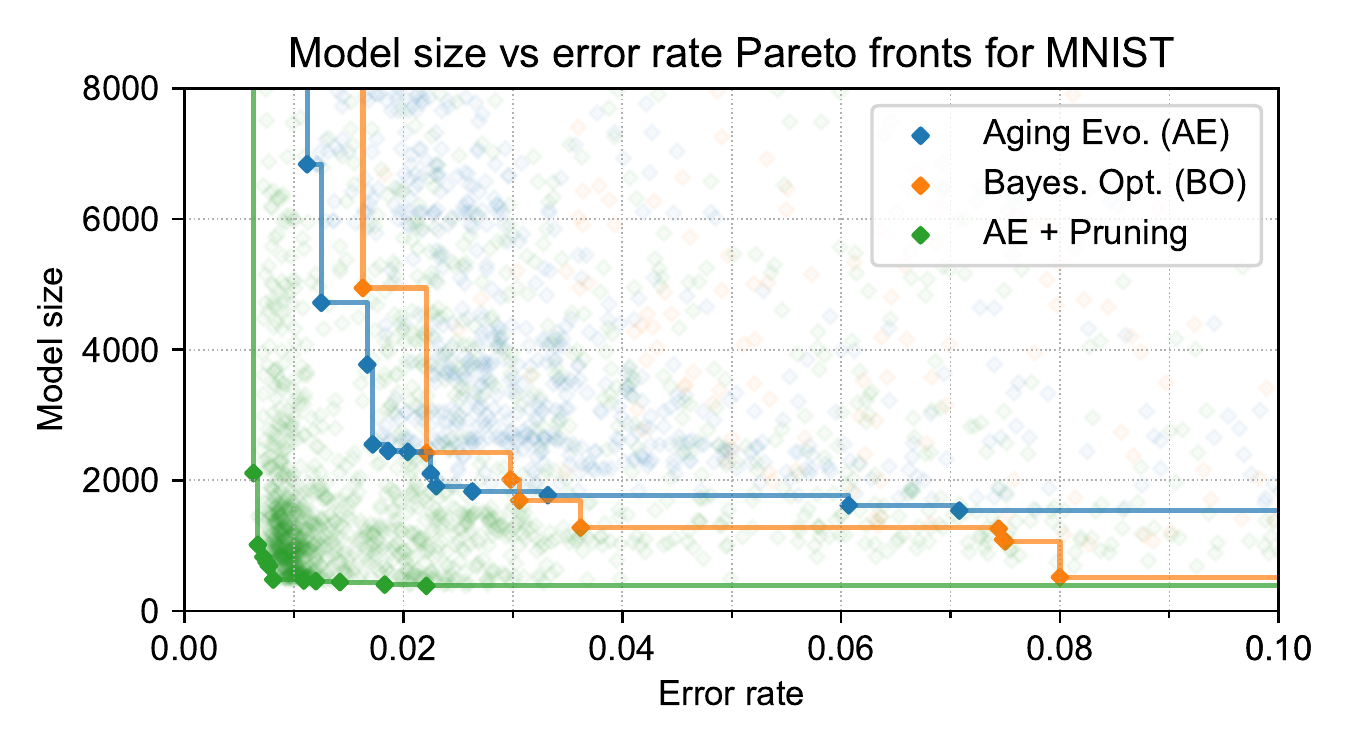}
    \caption{\textit{(Left)} error rate vs size of models discovered by μNAS on the Chars74K dataset; \textit{(Right)} ditto for the MNIST dataset. \textbf{The results show that μNAS configured with aging evolution and pruning finds the furthest advanced Pareto front.}}
    \label{fig:pareto_fronts}
\end{figure*}

The aim of the evaluation is twofold:
\begin{enumerate}
    \item \textbf{To show that all components of μNAS are required for obtaining small, performant models.} The design of μNAS is empirically validated through ablation studies. Namely, we address: 
    (a) whether aging evolution or Bayesian optimisation perform better at discovering the Pareto front;
    (b) whether including pruning yields more performant small models;
    (c) we check the utility of including both model size and peak memory usage objectives by searching without either.
    \item \textbf{To show that μNAS produces superior models}, as compared to previous work on CNNs for MCUs.
\end{enumerate}

\subsection{Datasets}
We evaluate μNAS on five image classification problems: MNIST~\cite{lecun2010mnist}, CIFAR-10~\cite{Krizhevsky09learningmultiple}, Chars74K~\cite{de2009character} (English subset: 26 uppercase and 26 lowercase letters + 10 digits), Fashion MNIST~\cite{xiao2017fashion} and Speech Commands~\cite{warden2018speech}. Datasets are split into training, validation and test sets, with the validation set used for evaluating objectives and tuning hyperparameters. Accuracy results are reported on the unseen test set. 

We also consider ``\textit{binary}'' versions of Chars74K and CIFAR-10, which have images partitioned into two subclasses. As far as we are aware, the full Chars74K or Fashion MNIST datasets do not have strong low resource usage baselines. Except for CIFAR-10, μNAS was run for 2000 steps. The search took up to 39 GPU-days for Speech Commands, and up to 3 GPU-days for the remaining tasks. We set constraints to either less than 64 KB of peak memory usage and storage, corresponding to our target MCUs, or less than 2KB, when required for comparison. Experiment details can be found in Appendix A. We also provide experiments on searching for sparse models (Appendix B).

\subsection{Determining the best μNAS configuration}

μNAS can use both AE and BO as a search algorithm and optionally use pruning during the search. We would like to determine which implemented configuration yields the best results for finding models with low resource usage and should therefore be used for further experiments. To do so, we run μNAS in three configurations on Chars74K and MNIST datasets: (1) plain aging evolution (AE); (2) plain Bayesian optimisation (BO); (3) AE with pruning. Figure~\ref{fig:pareto_fronts} shows the model size \emph{vs} error rate ($1.0 - \text{accuracy}$) of discovered models for each configuration. The lower-left corner of the plot contains models with low error and low resource usage; the further the Pareto front (highlighted in the plots) extends into that corner area, the better size vs accuracy trade-off μNAS is able to discover.

Search algorithm comparison shows that AE discovers a further advanced Pareto front than BO. We hypothesise that BO performs worse due to a smooth approximation to a network’s accuracy (GPs) being inadequate to capture the subtle differences between networks in the search space, especially when few points are available to build the surrogate in the initial stages of the search. 

We also observed that BO completes fewer search rounds compared to AE in the same allotted time. This is due to an added computational burden of (a) computing resource usage of many models when optimising an acquisition function, (b) updating the posterior model and (c) computing the kernel function after each new model has been trained.

The data shows that pruning is essential for finding models with low resource usage,  and AE with pruning outperforms other configurations implemented in μNAS. Here's why: model compression postulates that it is difficult to train small models alone from scratch. This is also confirmed by the inferior accuracy of models found by μNAS without pruning. Searching for larger base architectures to subsequently prune avoids training small models from scratch: a substantial number of channels/units are only pruned away after a model has reached a high level of generalisation. Therefore, AE with pruning (the search procedure summarised earlier in Section~\ref{sec:summary}) is the strongest and default design choice and will be used in further experiments.

\subsection{The utility of resource objectives}

\begin{figure*}[t]
    \centering
    \includegraphics[scale=0.59]{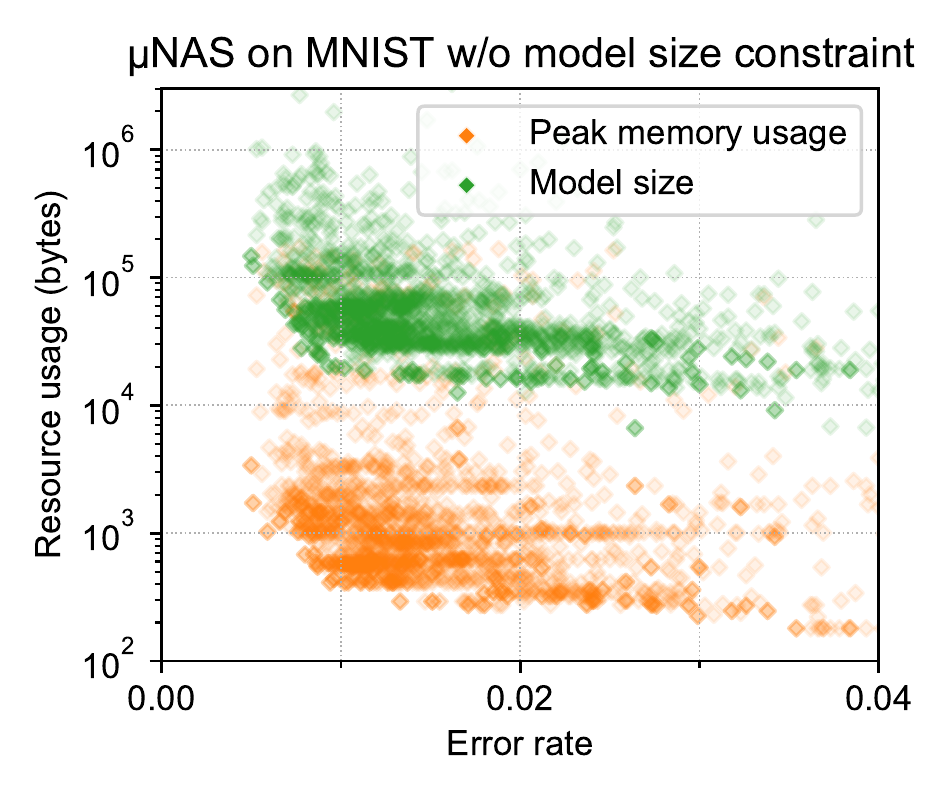}$\:$%
    \includegraphics[scale=0.59]{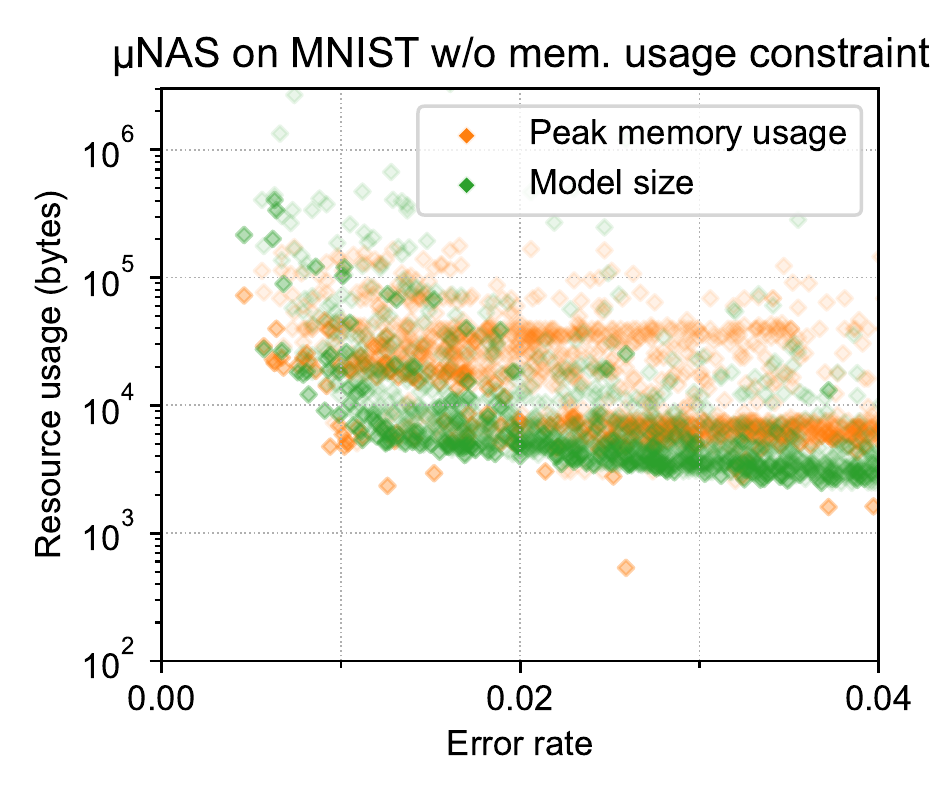}$\:$%
    \includegraphics[scale=0.59]{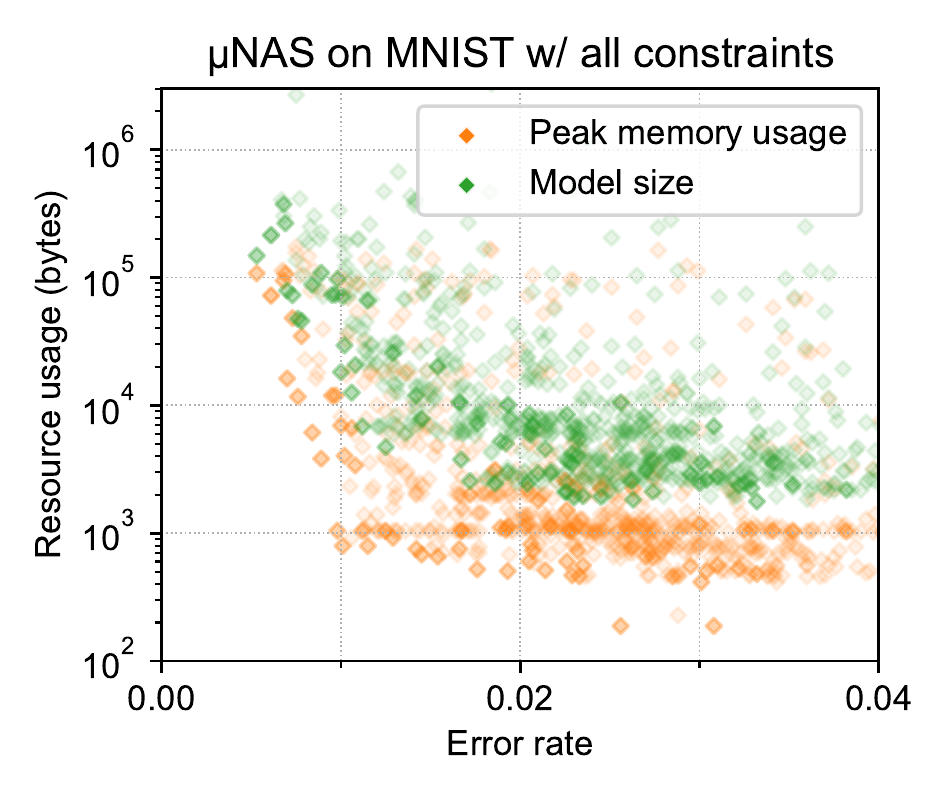}$\:$%
    \caption{Error rate vs resource usage properties of MNIST models; each architecture produces 2 data points: one for model size and one for peak memory usage. Latency (MACs) are omitted for clarity. μNAS configurations: \textit{(left)} model size was unconstrained, \textit{(middle)} peak memory usage was unconstrained, \textit{(right)} all constraints were present. When either constraint is absent, the other dominates the search; when both are present, a variety of trade-off points is discovered. \textbf{The plots show that both peak memory usage and model size objectives are influential and informative for the search.}}
    \label{fig:mnist_objectives}
\end{figure*}

Having developed precise resource usage computations, we would like to check whether including them actually provides meaningful input to the search by guiding it towards low resource usage models.

Here, we will observe how model size (MS) and peak memory usage (PMU) objectives influence the search. We run μNAS on MNIST in three modes: all constraints present, no model size (MS) constraint, and no peak memory usage (PMU) constraint. We expect these three regimes to exhibit a difference in the kinds of models found by μNAS: the PMU objective should drive the search towards narrow models (a lower memory bottleneck), MS---towards shallow models (fewer layers); when both are present, a trade-off between low resource usage and accuracy should be discovered. 

Found models are presented in Figure~\ref{fig:mnist_objectives}. We do not use pruning and set a generous latency constraint to avoid introducing any additional bias to the results (MACs are omitted from the plot). The data confirms that both objectives are needed for the search: (a) when MS is unconstrained, the search is dominated by the PMU constraint and driven to models with low memory usage; this also implies high accuracy, since it is possible to generate deep and narrow networks with high model size but a low memory bottleneck; (b) when PMU is unconstrained, the search is driven to models with low MS and average PMU; (c) when both objectives are present, μNAS discovers a variety of trade-off points between accuracy and resource usage, with models that have low to moderate PMU and MS.

\subsection{Discovered architectures}
Table~\ref{tab:unas_results} shows networks discovered by μNAS. For each task, we present a baseline model at multiple objective trade-off points, together with a model discovered by μNAS that either improves upon or closely match on all (known) metrics.

\newcommand{\CC}[1]{\cellcolor{lightergray}{#1}}
\begin{table*}[h]
    \centering
    \small
    \begin{tabular}{llrrrrr}
\toprule
\textbf{Dataset} & \textbf{Model} & \textbf{Acc. (\%)} & \textbf{Model size} & \textbf{RAM usage} & \textbf{MACs} & \textbf{Difference} \\
\midrule

MNIST 
& SpArSe~\cite{fedorov2019sparse}    & 98.64 & 2770 & $\geq$ 1960 B & \textit{unk.} &  Size $\uparrow 5.7{\times}$ \\
& SpArSe    & 96.49 & 1440 & $\geq$ 1330 B & \textit{unk.} & Acc.  $\downarrow 2.7\%$\\
& BonsaiOpt~\cite{kumar2017resource} & 94.38 & 490 & $<$ 2000 B & \textit{unk.} & Acc. $\downarrow 4.8\%$ \\
& ProtoNN~\cite{gupta2017protonn} & 95.88 & 63’900 & $<$ 64’000 B & \textit{unk.} & Acc. $\downarrow 3.3\%$ \\
\rowcolor{lightergray} \cellcolor{white} & \textit{μNAS (1174 steps, 1 GPU-day)} & {\bf 99.19} & \textbf{480} & \textbf{488 B} & 28.6 K & \\
\midrule

\multirow{2}{*}{\makecell[l] {CIFAR-10 \\ (binary) }}
& SpArSe    & 73.84 & 780 & $\geq$ 1280 B & \textit{unk.} & Acc. $\downarrow 3.7\%$ \\
& \CC{\textit{μNAS (1206 steps, 2 GPU-days)}} & \CC{{\bf 77.49}} & \CC{\textbf{685}} & \CC{\textbf{909 B}} & \CC{41.2 K} & \CC{} \\
\midrule

\multirow{2}{*}{\makecell[l] {Chars74K \\ (binary) }}
& SpArSe    & 77.78 & 460 & $\geq$\textbf{ 720 B} & \textit{unk.} & Acc. $\downarrow 3.4\%$ \\
& \CC{\textit{μNAS (743 steps, 0.5 GPU-day)}} & \CC{\bf 81.20} & \CC{\textbf{390}} & \CC{867 B} & \CC{107 K} & \CC{} \\
\midrule

\multirow{2}{*}{\makecell[l] {Speech \\ Commands }}
& RENA~\cite{zhou2018rena} & 94.04 & 47 K & \textit{unk.} & $\approx$700 M  & MACs $\uparrow 636{\times}$ \\
& RENA & {\bf 94.82} & 67 K & \textit{unk.} & $\approx$3’265 M & MACs $\uparrow 2968{\times}$\\
& DS-CNN~\cite{zhang2017hello} & 94.45 & {\bf $<$38.6 K} & \textit{unk.} & $\approx$2.7 M & MACs $\uparrow 2.2{\times}$ \\
& MCUNet~\cite{lin2020mcunet} & ${\approx}$91.20 & $<$ 1 M & 80 KB & \textit{unk.} & Acc. $\downarrow 4.4\%$ \\
& MCUNet & {\bf ${\approx}$95.91} & $<$ 1 M & 311 KB & \textit{unk.} & RAM $\uparrow 14.7{\times}$\\
& \CC{\textit{μNAS (1960 steps, 39 GPU-days)}} & \CC{\bf 95.58} & \CC{\textbf{37 K}} & \CC{\bf 21.1 KB} & \CC{\textbf{1.1 M}} & \CC{} \\
\cmidrule{2-7}

& MN-KWS (L)~\cite{banbury2020micronets} & \textbf{95.3\:\:} & 612 K & 208 K & \textit{unk.} & Size $\uparrow 31.8\times$ \\
\rowcolor{lightergray} \cellcolor{white} & \textit{μNAS (1514 steps, 30 GPU-days)} & \textbf{95.36} & \textbf{19.2 K} & 25.7 KB & {1.1 M} & \\
\midrule

\multirow{2}{*}{\makecell[l] {Fashion \\ MNIST}}
& reported~\cite{fashionmnistgh2020} & 92.50 & $\approx$100 K & \textit{unk.} & \textit{unk.} & Size $\uparrow 1.6{\times}$ \\
& \CC{\textit{μNAS (1161 steps, 3 GPU-days)}} & \CC{\bf 93.22} & \CC{\textbf{63.6 K}} & \CC{12.6 KB} & \CC{4.4 M} & \CC{} \\
\midrule

CIFAR-10
& LEMONADE~\cite{elsken2018efficient} & {\bf ${\approx}$91.77} & \textbf{10 K} & \textit{unk.} & \textit{unk.} \\
\rowcolor{lightergray} \cellcolor{white} & \textit{μNAS  (4205 steps, 23 GPU-days)} & 86.49 & 11.4 K & 15.4 KB & 384 K & Acc. $\downarrow 5\%$ \\
\midrule

\rowcolor{lightergray} \cellcolor{white} Chars74K
& \textit{μNAS (1755 steps, 2 GPU-days)} & 82.65 & 3.85 K & 9.75 KB & 279 K & \\
\rowcolor{lightergray} \cellcolor{white} & \textit{μNAS (1724 steps, 2 GPU-days)} & 76.05 & 13.9 K & 1.81 KB & 111 K & \\
\bottomrule
    \end{tabular}
    \caption{Pareto-optimal architectures discovered by μNAS vs others, together with search times. The last column compares a baseline to a model discovered by μNAS in the group (the last line of each group). ``\textit{Unknown (unk.)}'' denotes data not reported by the authors. \textbf{The results show that μNAS outperforms most baselines by either improving accuracy for comparable resource usage or vice versa.}}
    \label{tab:unas_results}
\end{table*}
 
The data shows that μNAS can discover truly tiny models, advancing the state-of-the-art for MCU deep learning for 6 out of 7 tasks considered. The models are trainable from scratch with pruning, and are within our assumed constraints of ``mid-tier'' MCUs with a 64 KB memory and storage limit.

The results show:
(a) an improved top-1 classification accuracy by up to 4.8\% for the same memory and/or model size footprint (see MNIST; CIFAR-10 (bin.), Chars74K (bin.); Speech Commands), or
(b) a reduced memory footprint by 4--13${\times}$ while preserving accuracy (see MNIST, Speech Commands, Fashion MNIST), or 
(c) a reduced number of MACs by at least ${\approx}$2${\times}$ (see Speech Commands).

Targeting Speech Commands in particular shows the importance of considering MACs during the search: even when model size and accuracy are comparable with the baselines, there is scope for discovering models that are orders of magnitude faster (in MACs). For MNIST, the search found a pruned (yet still dense) model with $<$ 0.5 KB parameters and $>$ 99\% accuracy. Earlier, Figure~\ref{fig:pareto_fronts} showed that μNAS discovered many models in this area of the accuracy-size trade-off for MNIST, confirming that this is not an outlier. 

For CIFAR-10, μNAS is able to outperform the baseline for the binary dataset; however, it falls short compared to the models discovered by LEMONADE~\cite{elsken2018efficient} for a full (10-class) dataset. Upon closer investigation, we find that the μNAS does gradually push the Pareto front, but requires many steps to do so (the search took 8000 steps, $\approx$40 GPU-days). We predict that if given more search time, μNAS can discover better models for CIFAR-10.

\section{Discussion}

We have both qualitatively and quantitatively justified the design decisions made in μNAS, showing that it is able to find accurate yet small models, suitable for deployment on microcontrollers. We now discuss overarching points relating to μNAS and how it can be modified in future work.

\subsection{Convergence of aging evolution (AE)} 
CIFAR-10 experiments showed that the search is slow at advancing the Pareto front. We found that this is due to the combined effect of the search space being highly granular and the search algorithm requiring many steps to converge. While considering hyperparameters and morphisms at a high granularity is instrumental for finding tiny models (see, for example, MNIST and other results), it also makes the search less efficient by requiring multiple steps to change any candidate model in the population significantly and training a model from scratch at each such step. Slow convergence is also likely to occur for large ImageNet-based image classification tasks, such as Visual Wake Words~\cite{chowdhery2019visual}, exacerbated by the fact that candidate models would also take longer to train (up to an hour) than on CIFAR-10. 

Thus, while using an evolutionary local search algorithm (like AE) on this search space yields small and performant models, the search can be time-consuming. We do, however, expect such searches to converge in time. Other NAS systems that use evolutionary optimisation with no model training shortcuts typically take many GPU-days: in fact, original AE experiments are reported to have taken over 3000 GPU-days~\cite{real2019regularized}. 

\ifpreprint
Aging evolution (AE) was chosen as a natural fit for navigating the highly granular search space through morphisms. We rejected other approaches in favour of AE due to it being evident that further development would be required:
\begin{itemize}
\item \textit{Gradient descent-based NAS.} In DARTS~\cite{liu2018darts} and derivative works, all layer options are instantiated to create a single model; however, we do not anticipate this to be feasible here due to the sheer number of connectivity options in the search space.
\item \textit{NAS with surrogate models.} Surrogate models, such as neural networks~\cite{tan2019mnasnet} or GPs have been used to drive the search or approximate a candidate network’s accuracy. However, we experimentally found that AE, which does not use a surrogate model, performed better than Bayesian Optimisation with GPs.
\end{itemize}
\fi

If search time is an issue, we envision the above being rectified by: (a) not using the same search space throughout the entire search process, for example, by using a parametrised space where granularity can vary throughout the search; or (b) guiding the search, such as by using ranking models~\cite{chau2020brp}; or (c) employing weight sharing to amortise the cost of training each candidate network.

\subsection{Reusability of μNAS components}
We demonstrated that having a granular search space and an accurate resource usage computation leads to a NAS that can discover performant MCU-sized networks. 

Overall, μNAS is modular: the search algorithm, objective scalarisation, the search space, the computation of resource usage and model compression (pruning) are all independent of each other and can be reused elsewhere or swapped out, for example, if a different network execution strategy is assumed or should a different search algorithm be needed. We make the source code of μNAS available publicly.

\section{Conclusions}
Neural architecture search is a powerful tool for automating model design, especially when manual design is challenging due to the need to balance high accuracy and fitting within extremely tight resource constraints. We showed that through the suitable design of the search space and explicit targeting of the three primary resource bottlenecks, we are able to create a NAS system, μNAS, that discovers resource-efficient models for a variety of image classification tasks. 


\bibliography{bibliography}

\begin{thebibliography}{47}
\providecommand{\natexlab}[1]{#1}
\providecommand{\url}[1]{\texttt{#1}}
\expandafter\ifx\csname urlstyle\endcsname\relax
  \providecommand{\doi}[1]{doi: #1}\else
  \providecommand{\doi}{doi: \begingroup \urlstyle{rm}\Url}\fi

\bibitem[{Apache TVM}(2020)]{mutvm2020}
{Apache TVM}.
\newblock {TinyML---How TVM is Taming Tiny}.
\newblock \url{https://tvm.apache.org/2020/06/04/tinyml-how-tvm-is-taming-tiny}
  (Accessed Aug 2020), 2020.

\bibitem[Banbury et~al.(2020)Banbury, Zhou, Fedorov, Navarro, Thakkar, Gope,
  Reddi, Mattina, and Whatmough]{banbury2020micronets}
Banbury, C., Zhou, C., Fedorov, I., Navarro, R.~M., Thakkar, U., Gope, D.,
  Reddi, V.~J., Mattina, M., and Whatmough, P.~N.
\newblock Micronets: Neural network architectures for deploying tinyml
  applications on commodity microcontrollers.
\newblock \emph{arXiv preprint arXiv:2010.11267}, 2020.

\bibitem[Cai et~al.(2018)Cai, Zhu, and Han]{cai2018proxylessnas}
Cai, H., Zhu, L., and Han, S.
\newblock {ProxylessNAS: Direct neural architecture search on target task and
  hardware}.
\newblock \emph{arXiv preprint arXiv:1812.00332}, 2018.

\bibitem[Chau et~al.(2020)Chau, Dudziak, Abdelfattah, Lee, Kim, and
  Lane]{chau2020brp}
Chau, T., Dudziak, {\L}., Abdelfattah, M.~S., Lee, R., Kim, H., and Lane, N.~D.
\newblock {BRP-NAS: Prediction-based NAS using GCNs}.
\newblock \emph{arXiv preprint arXiv:2007.08668}, 2020.

\bibitem[Cheng et~al.(2019)Cheng, Zhang, Yang, Yan, Li, Teague, Li, and
  Chen]{cheng2019swiftnet}
Cheng, H.-P., Zhang, T., Yang, Y., Yan, F., Li, S., Teague, H., Li, H., and
  Chen, Y.
\newblock {SwiftNet}: Using graph propagation as meta-knowledge to search
  highly representative neural architectures.
\newblock \emph{arXiv preprint arXiv:1906.08305}, 2019.

\bibitem[Cheng et~al.(2017)Cheng, Wang, Zhou, and Zhang]{cheng2017survey}
Cheng, Y., Wang, D., Zhou, P., and Zhang, T.
\newblock A survey of model compression and acceleration for deep neural
  networks.
\newblock \emph{arXiv preprint arXiv:1710.09282}, 2017.

\bibitem[Chowdhery et~al.(2019)Chowdhery, Warden, Shlens, Howard, and
  Rhodes]{chowdhery2019visual}
Chowdhery, A., Warden, P., Shlens, J., Howard, A., and Rhodes, R.
\newblock Visual wake words dataset.
\newblock \emph{arXiv preprint arXiv:1906.05721}, 2019.

\bibitem[David et~al.(2020)David, Duke, Jain, Reddi, Jeffries, Li, Kreeger,
  Nappier, Natraj, Regev, et~al.]{david2020tensorflow}
David, R., Duke, J., Jain, A., Reddi, V.~J., Jeffries, N., Li, J., Kreeger, N.,
  Nappier, I., Natraj, M., Regev, S., et~al.
\newblock Tensorflow lite micro: Embedded machine learning on tinyml systems.
\newblock \emph{arXiv preprint arXiv:2010.08678}, 2020.

\bibitem[De~Campos et~al.(2009)De~Campos, Babu, Varma, et~al.]{de2009character}
De~Campos, T.~E., Babu, B.~R., Varma, M., et~al.
\newblock Character recognition in natural images.
\newblock \emph{VISAPP (2)}, 7, 2009.

\bibitem[Elsken et~al.(2018)Elsken, Metzen, and Hutter]{elsken2018efficient}
Elsken, T., Metzen, J.~H., and Hutter, F.
\newblock Efficient multi-objective neural architecture search via {Lamarckian}
  evolution.
\newblock \emph{arXiv preprint arXiv:1804.09081}, 2018.

\bibitem[Elsken et~al.(2019)Elsken, Metzen, and Hutter]{elsken2019neural}
Elsken, T., Metzen, J.~H., and Hutter, F.
\newblock Neural architecture search: A survey.
\newblock \emph{Journal of Machine Learning Research}, 20\penalty0
  (55):\penalty0 1--21, 2019.

\bibitem[Falkner et~al.(2018)Falkner, Klein, and Hutter]{falkner2018bohb}
Falkner, S., Klein, A., and Hutter, F.
\newblock Bohb: Robust and efficient hyperparameter optimization at scale.
\newblock \emph{arXiv preprint arXiv:1807.01774}, 2018.

\bibitem[Fedorov et~al.(2019)Fedorov, Adams, Mattina, and
  Whatmough]{fedorov2019sparse}
Fedorov, I., Adams, R.~P., Mattina, M., and Whatmough, P.
\newblock {SpArSe}: Sparse architecture search for {CNNs} on
  resource-constrained microcontrollers.
\newblock In \emph{Advances in Neural Information Processing Systems}, pp.\
  4977--4989, 2019.

\bibitem[Fernandez-Marques et~al.(2020)Fernandez-Marques, Whatmough, Mundy, and
  Mattina]{fernandez2020searching}
Fernandez-Marques, J., Whatmough, P.~N., Mundy, A., and Mattina, M.
\newblock {Searching for Winograd-aware quantized networks}.
\newblock \emph{arXiv preprint arXiv:2002.10711}, 2020.

\bibitem[{Grand View Research}(2020)]{mcumarket2020}
{Grand View Research}.
\newblock {Microcontroller Market Size, Share \& Trends Analysis Report By
  Product (8-bit, 16-bit, 32-bit), By Application (Automotive, Consumer
  Electronics, Industrial, Medical Devices, Military \& Defense), And Segment
  Forecasts, 2020--2027}.
\newblock
  \url{https://www.grandviewresearch.com/industry-analysis/microcontroller-market}
  (Accessed Aug 2020), 2020.

\bibitem[Guo et~al.(2019)Guo, Zhang, Mu, Heng, Liu, Wei, and
  Sun]{guo2019single}
Guo, Z., Zhang, X., Mu, H., Heng, W., Liu, Z., Wei, Y., and Sun, J.
\newblock Single path one-shot neural architecture search with uniform
  sampling.
\newblock \emph{arXiv preprint arXiv:1904.00420}, 2019.

\bibitem[Gupta et~al.(2017)Gupta, Suggala, Goyal, Simhadri, Paranjape, Kumar,
  Goyal, Udupa, Varma, and Jain]{gupta2017protonn}
Gupta, C., Suggala, A.~S., Goyal, A., Simhadri, H.~V., Paranjape, B., Kumar,
  A., Goyal, S., Udupa, R., Varma, M., and Jain, P.
\newblock {ProtoNN}: Compressed and accurate knn for resource-scarce devices.
\newblock In \emph{International Conference on Machine Learning}, pp.\
  1331--1340, 2017.

\bibitem[He et~al.(2016)He, Zhang, Ren, and Sun]{he2016deep}
He, K., Zhang, X., Ren, S., and Sun, J.
\newblock Deep residual learning for image recognition.
\newblock In \emph{Proceedings of the IEEE conference on computer vision and
  pattern recognition}, pp.\  770--778, 2016.

\bibitem[Hsu et~al.(2018)Hsu, Chang, Liang, Chou, Liu, Chang, Pan, Chen, Wei,
  and Juan]{hsu2018monas}
Hsu, C.-H., Chang, S.-H., Liang, J.-H., Chou, H.-P., Liu, C.-H., Chang, S.-C.,
  Pan, J.-Y., Chen, Y.-T., Wei, W., and Juan, D.-C.
\newblock {MONAS: Multi-objective neural architecture search using
  reinforcement learning}.
\newblock \emph{arXiv preprint arXiv:1806.10332}, 2018.

\bibitem[Iandola et~al.(2016)Iandola, Han, Moskewicz, Ashraf, Dally, and
  Keutzer]{iandola2016squeezenet}
Iandola, F.~N., Han, S., Moskewicz, M.~W., Ashraf, K., Dally, W.~J., and
  Keutzer, K.
\newblock {SqueezeNet: AlexNet-level accuracy with 50x fewer parameters and $<$
  0.5 MB model size}.
\newblock \emph{arXiv preprint arXiv:1602.07360}, 2016.

\bibitem[Jacob et~al.(2018)Jacob, Kligys, Chen, Zhu, Tang, Howard, Adam, and
  Kalenichenko]{jacob2018quantization}
Jacob, B., Kligys, S., Chen, B., Zhu, M., Tang, M., Howard, A., Adam, H., and
  Kalenichenko, D.
\newblock Quantization and training of neural networks for efficient
  integer-arithmetic-only inference.
\newblock In \emph{Proceedings of the IEEE Conference on Computer Vision and
  Pattern Recognition}, pp.\  2704--2713, 2018.

\bibitem[Jin et~al.(2019)Jin, Song, and Hu]{jin2019auto}
Jin, H., Song, Q., and Hu, X.
\newblock {Auto-Keras}: An efficient neural architecture search system.
\newblock In \emph{Proceedings of the 25th ACM SIGKDD International Conference
  on Knowledge Discovery \& Data Mining}, pp.\  1946--1956, 2019.

\bibitem[Kandasamy et~al.(2018{\natexlab{a}})Kandasamy, Krishnamurthy,
  Schneider, and P{\'o}czos]{kandasamy2018parallelised}
Kandasamy, K., Krishnamurthy, A., Schneider, J., and P{\'o}czos, B.
\newblock {Parallelised Bayesian optimisation via Thompson sampling}.
\newblock In \emph{International Conference on Artificial Intelligence and
  Statistics}, pp.\  133--142, 2018{\natexlab{a}}.

\bibitem[Kandasamy et~al.(2018{\natexlab{b}})Kandasamy, Neiswanger, Schneider,
  Poczos, and Xing]{kandasamy2018neural}
Kandasamy, K., Neiswanger, W., Schneider, J., Poczos, B., and Xing, E.~P.
\newblock Neural architecture search with bayesian optimisation and optimal
  transport.
\newblock In \emph{Advances in neural information processing systems}, pp.\
  2016--2025, 2018{\natexlab{b}}.

\bibitem[Krizhevsky(2009)]{Krizhevsky09learningmultiple}
Krizhevsky, A.
\newblock Learning multiple layers of features from tiny images.
\newblock Technical report, 2009.

\bibitem[Kumar et~al.(2017)Kumar, Goyal, and Varma]{kumar2017resource}
Kumar, A., Goyal, S., and Varma, M.
\newblock Resource-efficient machine learning in 2 kb ram for the internet of
  things.
\newblock In \emph{International Conference on Machine Learning}, pp.\
  1935--1944, 2017.

\bibitem[LeCun \& Cortes(2010)LeCun and Cortes]{lecun2010mnist}
LeCun, Y. and Cortes, C.
\newblock {MNIST} handwritten digit database.
\newblock 2010.
\newblock URL \url{http://yann.lecun.com/exdb/mnist/}.

\bibitem[Liberis \& Lane(2019)Liberis and Lane]{liberis2019neural}
Liberis, E. and Lane, N.~D.
\newblock Neural networks on microcontrollers: saving memory at inference via
  operator reordering.
\newblock \emph{arXiv preprint arXiv:1910.05110}, 2019.

\bibitem[Lin et~al.(2020{\natexlab{a}})Lin, Chen, Lin, Cohn, Gan, and
  Han]{lin2020mcunet}
Lin, J., Chen, W.-M., Lin, Y., Cohn, J., Gan, C., and Han, S.
\newblock {MCUNet}: Tiny deep learning on iot devices.
\newblock \emph{arXiv preprint arXiv:2007.10319}, 2020{\natexlab{a}}.

\bibitem[Lin et~al.(2020{\natexlab{b}})Lin, Stich, Barba, Dmitriev, and
  Jaggi]{lin2020dynamic}
Lin, T., Stich, S.~U., Barba, L., Dmitriev, D., and Jaggi, M.
\newblock Dynamic model pruning with feedback.
\newblock \emph{arXiv preprint arXiv:2006.07253}, 2020{\natexlab{b}}.

\bibitem[Liu et~al.(2018)Liu, Simonyan, and Yang]{liu2018darts}
Liu, H., Simonyan, K., and Yang, Y.
\newblock {DARTS}: Differentiable architecture search.
\newblock \emph{arXiv preprint arXiv:1806.09055}, 2018.

\bibitem[Mei et~al.(2019)Mei, Li, Lian, Jin, Yang, Yuille, and
  Yang]{mei2019atomnas}
Mei, J., Li, Y., Lian, X., Jin, X., Yang, L., Yuille, A., and Yang, J.
\newblock {AtomNas}: Fine-grained end-to-end neural architecture search.
\newblock \emph{arXiv preprint arXiv:1912.09640}, 2019.

\bibitem[Mocerino \& Calimera(2019)Mocerino and Calimera]{mocerino2019coopnet}
Mocerino, L. and Calimera, A.
\newblock {CoopNet: Cooperative convolutional neural network for low-power
  MCUs}.
\newblock In \emph{2019 26th IEEE International Conference on Electronics,
  Circuits and Systems (ICECS)}, pp.\  414--417. IEEE, 2019.

\bibitem[Paria et~al.(2020)Paria, Kandasamy, and P{\'o}czos]{paria2020flexible}
Paria, B., Kandasamy, K., and P{\'o}czos, B.
\newblock A flexible framework for multi-objective bayesian optimization using
  random scalarizations.
\newblock In \emph{Uncertainty in Artificial Intelligence}, pp.\  766--776.
  PMLR, 2020.

\bibitem[Pham et~al.(2018)Pham, Guan, Zoph, Le, and Dean]{pmlr-v80-pham18a}
Pham, H., Guan, M., Zoph, B., Le, Q., and Dean, J.
\newblock Efficient neural architecture search via parameters sharing.
\newblock volume~80 of \emph{Proceedings of Machine Learning Research (PMLR)},
  pp.\  4095--4104, 2018.

\bibitem[Real et~al.(2019)Real, Aggarwal, Huang, and Le]{real2019regularized}
Real, E., Aggarwal, A., Huang, Y., and Le, Q.~V.
\newblock Regularized evolution for image classifier architecture search.
\newblock In \emph{Proceedings of the {AAAI} conference on artificial
  intelligence}, volume~33, pp.\  4780--4789, 2019.

\bibitem[Sandler et~al.(2018)Sandler, Howard, Zhu, Zhmoginov, and
  Chen]{sandler2018mobilenetv2}
Sandler, M., Howard, A., Zhu, M., Zhmoginov, A., and Chen, L.-C.
\newblock {MobileNet V2}: Inverted residuals and linear bottlenecks.
\newblock In \emph{Proceedings of the IEEE conference on computer vision and
  pattern recognition}, pp.\  4510--4520, 2018.

\bibitem[{STMicroelectronics}(2020)]{nucleoh743zi2020}
{STMicroelectronics}.
\newblock {STM32 Nucleo-144 development board with STM32H743ZI MCU, supports
  Arduino, ST Zio and morpho connectivity}.
\newblock \url{https://www.st.com/en/evaluation-tools/nucleo-h743zi.html}
  (Accessed Aug 2020), 2020.

\bibitem[Szegedy et~al.(2015)Szegedy, Liu, Jia, Sermanet, Reed, Anguelov,
  Erhan, Vanhoucke, and Rabinovich]{szegedy2015going}
Szegedy, C., Liu, W., Jia, Y., Sermanet, P., Reed, S., Anguelov, D., Erhan, D.,
  Vanhoucke, V., and Rabinovich, A.
\newblock Going deeper with convolutions.
\newblock In \emph{Proceedings of the IEEE conference on computer vision and
  pattern recognition}, pp.\  1--9, 2015.

\bibitem[Tan et~al.(2019)Tan, Chen, Pang, Vasudevan, Sandler, Howard, and
  Le]{tan2019mnasnet}
Tan, M., Chen, B., Pang, R., Vasudevan, V., Sandler, M., Howard, A., and Le,
  Q.~V.
\newblock {MNasNet}: Platform-aware neural architecture search for mobile.
\newblock In \emph{Proceedings of the IEEE Conference on Computer Vision and
  Pattern Recognition}, pp.\  2820--2828, 2019.

\bibitem[Warden(2018)]{warden2018speech}
Warden, P.
\newblock Speech commands: A dataset for limited-vocabulary speech recognition.
\newblock \emph{arXiv preprint arXiv:1804.03209}, 2018.

\bibitem[Xiao et~al.(2017)Xiao, Rasul, and Vollgraf]{xiao2017fashion}
Xiao, H., Rasul, K., and Vollgraf, R.
\newblock {Fashion-MNIST: a novel image dataset for benchmarking machine
  learning algorithms}.
\newblock \emph{arXiv preprint arXiv:1708.07747}, 2017.

\bibitem[Xie et~al.(2018)Xie, Zheng, Liu, and Lin]{xie2018snas}
Xie, S., Zheng, H., Liu, C., and Lin, L.
\newblock {SNAS: stochastic neural architecture search}.
\newblock \emph{arXiv preprint arXiv:1812.09926}, 2018.

\bibitem[{Zalando}(2020)]{fashionmnistgh2020}
{Zalando}.
\newblock {\texttt{zalandoresearch/fashion-mnist} Github repository}.
\newblock \url{https://github.com/zalandoresearch/fashion-mnist} (Accessed Aug
  2020), 2020.

\bibitem[Zhang et~al.(2017)Zhang, Suda, Lai, and Chandra]{zhang2017hello}
Zhang, Y., Suda, N., Lai, L., and Chandra, V.
\newblock {Hello Edge}: Keyword spotting on microcontrollers.
\newblock \emph{arXiv preprint arXiv:1711.07128}, 2017.

\bibitem[Zhou et~al.(2018)Zhou, Ebrahimi, Ar{\i}k, Yu, Liu, and
  Diamos]{zhou2018rena}
Zhou, Y., Ebrahimi, S., Ar{\i}k, S.~{\"O}., Yu, H., Liu, H., and Diamos, G.
\newblock Resource-efficient neural architect.
\newblock \emph{arXiv preprint arXiv:1806.07912}, 2018.

\bibitem[Zoph \& Le(2016)Zoph and Le]{zoph2016neural}
Zoph, B. and Le, Q.~V.
\newblock Neural architecture search with reinforcement learning.
\newblock \emph{arXiv preprint arXiv:1611.01578}, 2016.

\end{thebibliography}
\bibliographystyle{mlsys2020}


\newpage
\clearpage

\appendix
\section{Experiment configuration}
\subsection{Resource bound configurations}
In μNAS, we control which area of the Pareto front gets explored by specifying a preferred trade-off between objectives. The objective scalarisation (Equation~\ref{eqn:scalarised_goal}) together with coefficient sampling force the search to highly penalise any objectives that exceed their specified bound, and not preferentially penalise any objective that is within its bounds. We give our requested bounds in Table~\ref{tab:unas_bounds}.

\begin{table}[h]
    \centering
    \begin{tabular}{rrrr}
    \toprule
    \multicolumn{4}{c}{\bf Requested objective bounds} \\
    \midrule
    \textbf{Error} & \textbf{Peak memory} & \textbf{Model size} & \textbf{MACs} \\
    \midrule
    \multicolumn{4}{l}{\textit{MNIST}} \\ $<$ 0.0350 & $<$ 2.5 KB & $<$ 4.5 K & $<$ 30 M \\
    \midrule
    \multicolumn{4}{l}{\textit{Chars74K}} \\ $<$ 0.3000 & $<$ 10 KB & $<$ 20 K & $<$ 1 M \\
    \midrule
    \multicolumn{4}{l}{\textit{CIFAR-10}} \\ $<$ 0.1800 & $<$ 75 KB & $<$ 75 K & $<$ 30 M \\
    \midrule
    \multicolumn{4}{l}{\textit{Speech Commands}} \\ $<$ 0.0850 & $<$ 60 KB & $<$ 40 K & $<$ 20 M \\
    \midrule
    \multicolumn{4}{l}{\textit{Fashion MNIST}} \\ $<$ 0.1000 & $<$ 64 KB & $<$ 64 K & $<$ 30 M \\
    \bottomrule
    \end{tabular}
    \caption{Resource bounds requested from μNAS for each dataset.}
    \label{tab:unas_bounds}
\end{table}

\subsection{Dataset preprocessing and model training schedule}
\footnotetext{\url{https://github.com/tensorflow/tensorflow/blob/master/tensorflow/examples/speech_commands/input_data.py}}
Dataset preprocessing, as well as model training and pruning information, is given in Table~\ref{tab:unas_training_conf} (see next page). 

\textbf{Additional notes:} Models CIFAR-10 and Fashion MNIST also have a Dropout layer w/ rate = 0.15 inserted before every fully-connected layer, except the last one. The models have been quantised for comparison against baselines in Table~\ref{tab:unas_results} after training and, like \citet{lin2020mcunet}, we observe no loss in accuracy. This is not surprising due to the full-integer quantisation in TensorFlow Lite~\cite{jacob2018quantization} being very expressive: at a cost of extra operations, it allows to represent any finite range in 256 steps (for 8-bit quantisation), and weight decay used during encourages model weights to stay within reasonable bounds.

\section{Sparse models}
Sparse neural networks allow for an even more significant reduction of model size without sacrificing accuracy. To compare our models with the \textit{SpArSe} NAS~\cite{fedorov2019sparse}, we also allow μNAS to perform unstructured pruning using the same pruning method described in Section~\ref{sec:model_compression}, which results in models with sparse weight matrices. As opposed to dense operations which had been assumed so far, sparse models are executed using sparse matrix multiplications. Thus previously considered latency and peak memory usage constraints would no longer be accurate and are not included.

μNAS search results for sparse models are presented in Table~\ref{tab:sparse_models}. Results show an increased accuracy delivered by μNAS for a comparable model size (up to 4.6\%), or a smaller model size (up to two-fold) for comparable accuracy, depending on the task.

\begin{table}[]
    \centering
    \begin{tabular}{llrr}
    \toprule
    \textbf{Dataset} & \textbf{Model} & \textbf{Accuracy (\%)} & \textbf{Size} \\
    \midrule
    Chars74K & SpArSe & 67.82 & 1.7 K \\
    & \textit{μNAS (ours)} & \textbf{72.40} & 1.62 K \\
    \midrule
    \multirow{2}{*}{\makecell[l]{CIFAR-10 \\ (binary)}} & SpArSe & 81.77 & 3.2 K \\
    & \textit{μNAS (ours)} & \textbf{82.69} & \textbf{2.23 K} \\
    \cmidrule{2-4}
    & SpArSe & 76.66 & 1.4 K \\
    & \textit{μNAS (ours)} & 76.72 & \textbf{0.94 K} \\
    \midrule
    MNIST & SpArSe & 99.16 & 1 K \\
    & \textit{μNAS (ours)} & 99.19 (dense) & 480 \\
    \bottomrule
    \end{tabular}
    \caption{Sparse models found by μNAS vs others. \textbf{Results show an increased accuracy delivered by μNAS for a comparable model size, or a smaller model size for comparable accuracy.}}
    \label{tab:sparse_models}
\end{table}

\begin{table*}[p]
    \centering
    \begin{tabular}{p{1.5cm}p{6cm}p{4cm}p{3cm}}
    \toprule
    \textbf{Dataset} & \textbf{\makecell[l]{Data preprocessing \\ and augmentation}} & \textbf{\makecell[l]{Training and \\ optimiser configuration}} & \textbf{\makecell[l]{Pruning \\ configuration}} \\
    \midrule
    MNIST & \makecell[l]{
    \ipoint random rotate by +/- 0.2 rad with p = 0.3; \\
    \ipoint random shift (2, 2); \\
    \ipoint random flip L/R} & \makecell[l]{
    SGDW: \\
    \ipoint learning rate = 0.005, \\
    \ipoint momentum = 0.9, \\
    \ipoint weight decay = 4e-5; \\
    \ipoint epochs = 30; \\
    \ipoint batch size = 128.} & \makecell[l]{
    \ipoint target sparsity \\ in [0.05; 0.80]; \\
    \ipoint pruning between \\ epochs 3 and 18;} \\
    \midrule
    Chars74K & \makecell[l]{
    \ipoint image size 48x48 (32x32 for binary) \\
    \ipoint random split into 5000, 705, 2000 \\
    \ipoint images for train, val. \& test sets; \\
    \ipoint random shift by $\pm$ 10\% of H/W
    } & \makecell[l]{
    SGDW: \\
    \ipoint learning rate = 0.01, \\ 0.005 from epoch 35, \\ 
    \ipoint momentum = 0.9, \\
    \ipoint weight decay = 1e-4, \\
    \ipoint epochs = 60, \\
    \ipoint batch size = 80.} & \makecell[l]{
    \ipoint target sparsity \\ in [0.10; 0.85]; \\
    \ipoint pruning between \\ epochs 20 and 53} \\
    \midrule
    CIFAR-10 & \makecell[l]{
    \ipoint normalisation; \\
    \ipoint random flip L/R; \\
    \ipoint random shift (4, 4);
    } & \makecell[l]{
    SGDW: \\ 
    \ipoint learning rate = 0.01, \\ 0.005 from epoch 35, \\ 0.001 from epoch 65, \\
    \ipoint momentum = 0.9, \\
    \ipoint weight decay = 1e-5, \\ 
    \ipoint batch size = 128, \\
    \ipoint epochs = 80.
    } & \makecell[l]{
    \ipoint target sparsity \\ in [0.10; 0.90]; \\
    \ipoint pruning between \\ epochs 30 and 60;
    } \\
    \midrule
    {\makecell[l]{Speech \\ Commands}} &
    \makecell[l]{as given {here}\protect\footnotemark, with 30ms window size, \\ 10ms stride, and MFCCs extracted between \\ 20 Hz and 4000 Hz.
    } &
    \makecell[l]{
    AdamW: \\
    \ipoint learning rate = 0.0005, \\ 0.0001 from epoch 20, \\ 2e-5 from epoch 40; \\
    \ipoint weight decay = 1e-5; \\
    \ipoint batch size = 50; \\
    \ipoint epochs = 45.
    } & \makecell[l]{
    \ipoint target sparsity \\ in [0.10; 0.90]; \\
    \ipoint pruning between \\ epochs 20 and 40
    } \\
    \midrule
    \makecell[l]{Fashion \\ MNIST} & \makecell[l]{
    \ipoint random rotate by +/- 0.2 rad with p = 0.3; \\
    \ipoint random shift (2, 2); \\
    \ipoint random flip L/R } & \makecell[l]{
    SGDW:\\
    \ipoint learning rate = 0.01, \\ 
    0.005 from epoch 20, \\ 0.001 from epoch 35; \\ 
    \ipoint momentum = 0.9, \\
    \ipoint weight decay = 1e-5, \\
    \ipoint epochs = 45; \\
    \ipoint batch size = 128.
    } & \makecell[l] {
    \ipoint target sparsity \\ in [0.05; 0.90]; \\
    \ipoint pruning between \\ epochs 3 and 38
    }\\
    \bottomrule
    \end{tabular}
    \caption{Dataset preprocessing, model training and pruning configurations used in our experiments.}
    \label{tab:unas_training_conf}
\end{table*}


\end{document}